\documentclass[10pt,twocolumn]{article} 
\usepackage{simpleConference}
\usepackage{times}
\usepackage{graphicx}
\usepackage{amssymb}
\usepackage{url,hyperref}
\RequirePackage[natbibapa]{apacite}

\begin{document}

\title{Multi-site Diagnostic Classification Of Schizophrenia Using 3D CNN On Aggregated Task-based fMRI Data}

\author{Vigneshwaran S, V Bhaskaran \\
Sri Sathya Sai Institute of Higher Learning, India \\
\{vigneshwaranpersonal@gmail.com, vbhaskaran@sssihl.edu.in \}
}

\maketitle
\thispagestyle{empty}

\begin{abstract}
In spite of years of research, the mechanisms that underlie the development of schizophrenia, as well as its relapse, symptomatology, and treatment, continue to be a mystery. The absence of appropriate analytic tools to deal with the variable and complicated nature of schizophrenia may be one of the factors that contribute to the development of this disorder. Deep learning is a subfield of artificial intelligence that was inspired by the nervous system. In recent years, deep learning has made it easier to model and analyse complicated, high-dimensional, and nonlinear systems. Research on schizophrenia is one of the many areas of study that has been revolutionised as a result of the outstanding accuracy that deep learning algorithms have demonstrated in classification and prediction tasks. Deep learning has the potential to become a powerful tool for understanding the mechanisms that are at the root of schizophrenia. In addition, a growing variety of techniques aimed at improving model interpretability and causal reasoning are contributing to this trend. Using multi-site fMRI data and a variety of deep learning approaches, this study seeks to identify different types of schizophrenia. Our proposed method of temporal aggregation of the 4D fMRI data outperforms existing work. In addition, this study aims to shed light on the strength of connections between various brain areas in schizophrenia individuals.
\end{abstract}

\section{Introduction}
The application of functional magnetic resonance imaging (fMRI) to research human brain function and disorders has evolved as a strong neuroimaging method. It is one of the most crucial methods for determining how brain areas communicate with one another to complete certain tasks. It offers a potential method for studying the interaction between geographically distant separate brain areas that are engaged in a task at the same time.

The blood oxygenated level-dependent (BOLD) signal in fMRI is not random, but rather temporally coherent between geographically distant functionally consistent areas. Functional brain networks are formed by functionally connected areas \citep{1, 2}. The identification of functional activation in fMRI data during task or task-free "resting state" is crucial in neuroscience for pre-surgical planning and the diagnosis of neuropsychiatric diseases such as Alzheimer's \citep{3}, autism \citep{4}, and schizophrenia \citep{5}.

Several ways have been tested to diagnose diseases or classify persons as normal or ill. In general, the low SNR of fMRI data presents difficulties in using this data to diagnose neuropsychiatric disorders. To the best of our knowledge, no standard automated fMRI instrument is available in hospitals for illness diagnosis. This has prompted researchers to investigate the development of such systems that can assist doctors. Researchers, for example, have used stationary functional connectivity (FC) as a biomarker to differentiate patients with neurological and psychiatric diseases such as Alzheimer's \citep{6} or Schizophrenia from normal subjects \citep{7} because FC has been discovered to be changed in several neuropsychological diseases \citep{3, 8, 9, 10}.

In this study, we first collected fMRI raw datasets from two publicly available repositories including 300+ participants and developed a 3D convolutional network to train on the images by aggregating information over the timesteps for every subject for the automatic diagnosis of individuals with schizophrenia. We compare our model with various baseline models and record the performance.

\section{Related Work}

Machine learning has had a significant impact on schizophrenia diagnosis studies. Along with MRI and fMRI, the researchers analysed data from genetics, electroencephalography, and even audio interviews \citep{11}. A combination of fMRI and genomics data from a single location was utilized to perform canonical correlation analysis using two fully linked, sparse autoencoders followed by SVM \citep{12}. There were also several studies that combined fMRI and structural MRI; the fMRI tasks included resting state, letter-n back task, auditory oddball, and audiovisual stimuli tasks that elicited negative and neutral emotion \citep{13, 14, 15, 16, 17}. These studies used a variety of strategies to differentiate between classes, including 3D activation maps, functional connectomes, ICA decomposition, and independent polygenic risk scores along with modelling techniques such as artificial neural networks and decision tree classifiers.

Our research was inspired by a publication in which the authors attempted to discover consistent patterns and connections between brain areas and also categorized participants using a variety of machine learning models \citep{18}. However, their study was focused on a single site, and we desired to incorporate data from numerous locations, which is where \citep{19} came in handy. The writers of this article combined private and publicly available datasets. That, we believe, is the only effort in this area that utilized such a vast sample space for model training. We were unable to locate many open fMRI datasets, particularly those that included both T1 weighted and resting-state fMRI. Finally, we chose two locations, COBRE and UCLA.

\section{Materials and Methods}

\subsection{Participants}
The dataset is made up of 120 schizophrenia patients and 206 healthy people who were gathered from two different imaging resources. Refer to Table 1. The first subgroup was developed at the University of California, Los Angeles (UCLA: 50 schizophrenics and 122 healthy controls; patients were allowed to use stable medications) \citep{20}. The second group comes from the Center for Biomedical Research Excellence (COBRE) (COBRE: 69 Schizophrenia Strict Spectrum patients, 11 Schizoaffective Spectrum patients, and 84 healthy controls; all of the patients were on antipsychotic medications) \citep{21}.

All individuals were screened and rejected if they had a history of neurological illness, mental retardation, serious head trauma resulting in a loss of consciousness lasting more than 5 minutes, or a history of substance addiction or dependency within the previous 12 months. The Structured Clinical Interview for DSM-IV Disorders was utilized to obtain diagnostic information (SCID).

\begin{table*}[h]
	\begin{tabular}{l l l l l l l l l }
	\hline
	Study    & Age(mean ${\pm}$ std) & Age     & M   & F   & Control & Shcz Strict & Schz Affective & Schizophrenia \\ \hline
	UCLA     & 33.00 ${\pm}$ 9.07    & 21 - 50 & 103 & 69  & 122     & 0           & 0              & 50            \\
	COBRE    & 38.28 ${\pm}$ 12.60   & 18 - 66 & 125 & 39  & 84      & 69          & 11             & 0             \\
	Combined & 33.58 ${\pm}$ 11.24   & 18 - 66 & 228 & 108 & 206     & 69          & 11             & 50            \\ \hline
	\end{tabular}
	
	\caption{Demographics of two sites of data }
\end{table*}

\subsection{Image Acquisition}
On a Siemens Erlangen 3.0 Tesla Trim Trio scanner, MR images from both locations were collected. There were, however, modest changes in the methods used to capture the photos. The COBRE dataset was collected in complete k-space EPI sequences in a single shot. The time repetitions were set to 2000 milliseconds, and the echo time was set to 29 milliseconds. With a field of vision of 192 mm and a thickness of 4 mm, there is no gap between the 32 slices. A resting state run was performed on each participant, yielding 150-time steps. An asymmetrical spin-echo echo-planar sequence was used to collect the UCLA dataset. The time repetitions were set to 2000 milliseconds, and the echo time was set to 30 milliseconds. With a field of vision of 192 mm and a thickness of 4 mm, there is no gap between the 34 slices. A resting state run was performed on each participant, yielding 152-time steps.

\subsection{Data Preprocessing}
We performed an extensive 7-step pipeline to preprocess the resting-state fMRI data. It has been recorded by research that enhanced preprocessing of the fMRI data improves the performance of modelling \citep{22}. First, we apply slice-time correction to each voxel's time series using SPM known as Hanning-Windowed Sinc Interpolation (HWSI). Afterwards, we subtract 7 timesteps from the beginning of each subject to account for magnetic saturation. Using FSL-MCFLIRT \citep{23}, we were able to correct the fMRI for motion artefacts induced by low-frequency drifts, which might have a detrimental influence on the time course decomposition of the data. We eliminated the skull and neck voxels from the structural T1 weighted image using the FSL-BET \citep{24} corresponding to each fMRI time series from the structural T1 weighted image. As a result, we coregistered the structural image with the functional image, and the process was completed. Following that, we matched the registered brains to the Montreal Neurological Institute standard 3mm brain template (MNI152) \citep{25}. Using a Gaussian kernel with a full-width half-maximum of 4mm, spatial smoothing was performed to a time series of data (FWHM). The final step was to downscale the images to the integer format in order to decrease the amount of disc space consumed.

\subsection{Functional Connectivity Measure}
In order to extract the functional connectomes, we used two brain atlases for two different purposes. One was to use it for classification purposes whereas the other one is for understanding brain connectivity to understand which regions of interest are the biomarkers for schizophrenia. For the purpose of training the classification model, we leaned towards using the Harvard-Oxford cortical parcellation \citep{26} sampled at 2mm with a threshold of a probability of 0.25. For every subject, we initially computed the time series for each of the ROIs provided by the atlas by standardizing and transforming the original fMRI space into the atlas space without performing confounds regression. Following this, we were able to compute the Pearson correlation coefficient between the voxels over the time step. Finally, we had a 96 X 96-shaped correlation matrix for each of the subjects. The next brain atlas we used is MSDL \citep{27} which has 39 regions of interest. We repeated the same procedures on this atlas as well. Once we computed the functional connectomes, we used those to visualize the results.

\begin{figure}[h]
\includegraphics[width=\linewidth]{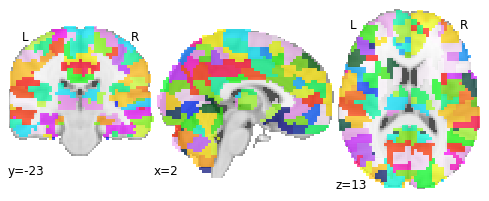}
\caption{ROI plot of Harvard-Oxford cortical parcellation atlas}
\label{fig:follow_half}
\end{figure}

\subsection{Random Forest}
Functional connectomes were first flattened to make it into a 1D vector of length 9216. To find the best hyperparameters, a grid search was performed. The best parameter turned out to be an entropy-based loss, with a max depth of 6, maximum features of 96 at every node, and 500 estimators. As the dataset is comparatively small, the classifier overfitted the training set. The performance and metrics are discussed in the results section.

\subsection{Vanilla Neural Network}
The input to the model is a flattened connectome matrix just like the random forest input. We tried to keep the model as simple as possible. We started with 213 neurons and gradually reduced it to 28 neurons before having a final layer with a single unit. All the hidden layers were activated using ReLU functions. As the problem is a binary classification, the last layer is a sigmoid-activated neuron. We played with various dropout strategies before settling with 30 \% between the layers.

\subsection{Sparse Autoencoder}
The paper \citep{19} we took inspiration from the multi-site approach and used sparse autoencoders with discriminant features. We tried to replicate the work with the two-site data pool to which we had access. We again flattened the connectomes to be passed as inputs. Our latent representation was of dimension 1024 which was reduced from 9216. We used 4096 and 2048 units respectively in two layers between the input and latent representation. We observed that SeLU worked better than ReLU in retaining information. We added l1 regularization with a penalty term of 0.001. We used a stochastic gradient descent optimizer with a learning rate of 0.1 and trained for 30 epochs.

Then we used the encoder to reduce the data of all the training samples and trained a simple vanilla neural network to classify. This network had 4 layers of neurons(512, 128, 64, 32) with ReLU activations and a final output layer with sigmoid activation. We followed the same drop strategy as in the Vanilla Neural Network model. We used an Adam optimizer as it was generalizing better.

\subsection{3D CNN}
Majority of the neuroimaging research that involves task-based fMRI data for classification use some method to transform into either functional connectomes or other forms of dimensionality reduction. Analysing a time series of 3D vectors is computationally expensive and there are very limited available methods that try to capture the temporal cohesion. These methods come from different domains and have their own set of challenges when we try to adapt them to neuroimaging problems. We still wanted to analyse the images. So instead of processing the 4D data, we tried to capture the information in a 3D plane. It is recorded in multiple pieces of research that schizophrenia is caused due to reduced neuronal activity. So we aggregated the voxel values to try and record either maximal values or minimal values. The intuition behind this is that the network might be able to observe reduced patterns in certain parts of the brain.

\begin{figure}[h]
\includegraphics[width=\linewidth]{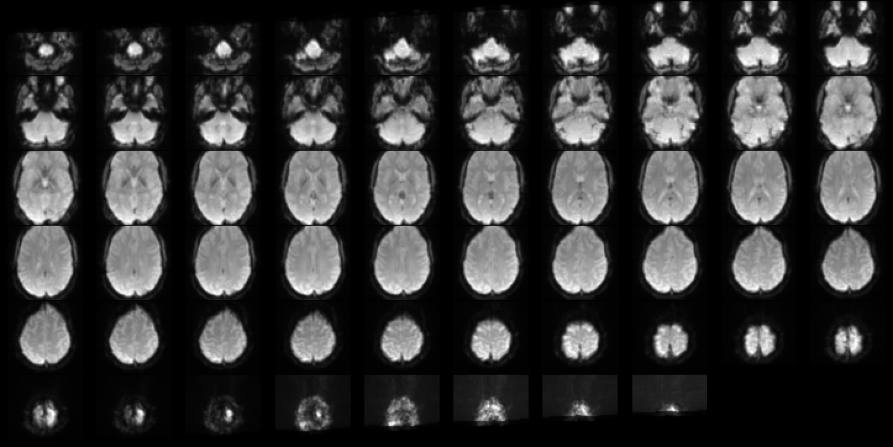}
\caption{A single subject aggregated over the $4^{th}$ axis}
\label{fig:follow_half}
\end{figure}

We first started by aggregating data over the 4th axis by their maximum values and minimum values for all the subjects. So we reduced the data size by many folds and captured the data in a 3D space. Now we had a lot of techniques that could be applied to 3D data. A lot of work has been performed in the domain of CT scans. So we adopted a 3D convolutional neural network which was commonly used to classify lung images with pneumonia. We performed minimal argumentation by rotating the 3D image with some angle. The random angle was selected from (-20, -10, -5, 5, 10, 20) every time it was augmented.

\begin{figure}[h]
\includegraphics[width=\linewidth]{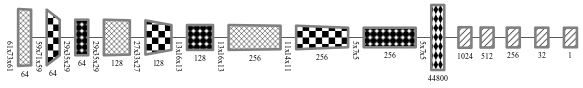}
\caption{Architecture of 3D CNN model}
\label{fig:follow_half}
\end{figure}

The input is a 61 X 73 X 61 matrix. A series of 3D convolutions was applied with 64, 128, and 256 filter sizes respectively all with a kernel size of 3 with ReLU activation function before passing it through fully connected layers. In between the convolutions layers, we performed max pooling with a size of 2 followed by batch normalization to account for overfitting. See Figure 3 for a detailed representation.

The performance was better than what was recorded in \citep{18}. The best model of that paper, which was a linear support vector machine-scored 74\% in terms of accuracy. Even though our maximum aggregation was giving this performance, the minimum aggregation did not perform well and scored an accuracy score of 66\%.

\section{Results}

\subsection{Model Performance}
The random forest model on the connectome matrice scored an accuracy of 69.5\%. The sensitivity and specificity were 0.86 and 0.38 respectively. There was no difference in terms of model performance in both the vanilla neural network and also the sparse autoencoder. Both of them recorded an accuracy of 71\% which is greater than the random forest. The type 1 errors and type2 errors were also improved in terms of ratio as the sensitivity and specificity were 0.77 and 0.62 respectively. Finally, our proposed methods of aggregated time series scored an accuracy of 77\% which is significantly better than the ones reported in \citep{18}. The sensitivity was great as it was 0.91 and the specificity metric was 0.54. Figure 4 has better visualization of the model performance metrics.

\begin{figure}[h]
\includegraphics[width=\linewidth]{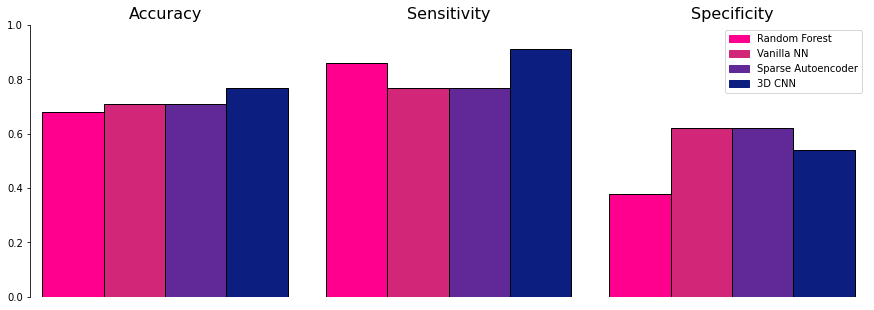}
\caption{Performance Metrics}
\label{fig:follow_half}
\end{figure}

\subsection{Brain Connectivity}
As the objective of this study was to also provide explainability that would result in a better understanding of the disorder, we wanted to take a deeper look at the interactions between the brain regions between the healthy subjects and schizophrenia patients. So we started by taking the functional connectomes of both the groups and averaging them. This way, now it was possible to have a single connectome that captured the variance of the entire group. This functional connectome was used to perform a connectivity plot.

Figure 5 was rendered computed using the MSDL atlas that we discussed in the functional connectivity measure section. At first sight, we could not make out much of a difference by looking at it. Therefore, this matrix was used to visualize a brain mask to bring out the connectivity between regions. We also visualized the connection in 3D space but due to the confines of this 2D document, it isn't recorded in this paper. However, the 3D visualizations are available in the repository.

\begin{figure}[h]
\includegraphics[width=\linewidth]{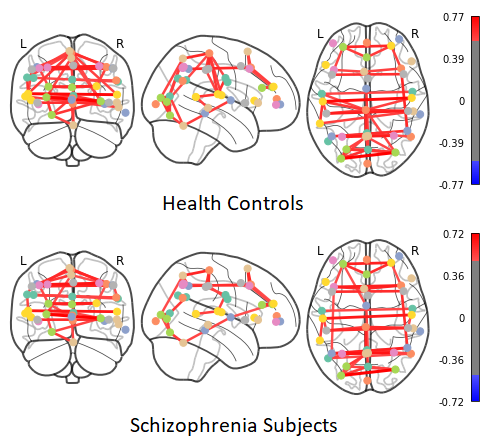}
\caption{Brain connectivity of health controls and schizophrenic subjects}
\label{fig:follow_half}
\end{figure}

\section{Discussion}

According to the findings of this study, a deep learning approach for reliable and early identification of Schizophrenia illness using brain scans has been proposed. When it comes to distinguishing schizophrenic individuals from normal people, the suggested approach uses a modified version of the Sparse Autoencoder algorithm in conjunction with a 3D Convolutional Neural Network. This aids in the correct and timely diagnosis of schizophrenia, as well as the early intervention necessary to combat the disorder. The suggested approach solves the limits of manual clinical elements as well as the limitations of existing procedures, which were discussed in the preceding sections.

The suggested approach primarily contributes to improving the accuracy with which Schizophrenia illness may be detected using brain scans such as functional magnetic resonance imaging (fMRI). As a result, it may be used to aid in the diagnosis of Schizophrenia illness using various modalities of brain imaging, and it can also be used as a CAD for the diagnosis of the disorder.

\subsection{Future Scope}
The work reported in this study is a first step toward directly exploiting image data for classification purposes (i.e., converting 4D to 3D space). The poor results obtained indicate that more investigation is required to fully comprehend this issue. The next step in our study in this sector is to explicitly incorporate the many kinds of noise known to corrupt fMRI results, such as magnetic field inhomogeneity or temporal signal drift. These parameters can be determined using phantoms. In a second strategy, classifiers for a single scanning site would be developed, but they would be trained to utilize priors learned from the other scanning sites as priors instead. Several organizations are already looking at the application of multitask learning to achieve this aim, and their first findings indicate that this would be a good option to follow in the future. Significant advancements in the field of 3D environment segmentation would benefit the neuroimaging community significantly since concepts might be borrowed from the field. A study of graph algorithms on connectome brain graphs may reveal insights to detect schizophrenia or provide explainability.

\subsection{Conflicts of interest}
The authors report no conflicts of interest.

\subsection{Supplementary Data}
We have made all the code, resources, and side notes required to reproduce the work publicly available at this GitHub repository: https://github.com/s-vigneshwaran/Schizophrenia-Classification-from-multi-site-fMRI-data.

\bibliographystyle{apacite}
\bibliography{refs}
\end{document}